\documentclass{article}
\usepackage{spconf,amsmath,graphicx}
\usepackage{times}
\usepackage{epsfig}
\usepackage{graphicx}
\usepackage{amsmath}
\usepackage{amssymb}
\usepackage{subfig}
\usepackage[hyphens]{url}
\newcommand{\etal}{\textit{~et~al.~}}

\title{Face Authentication from Grayscale Coded Light Field}
\name{Dana~Weitzner,~David~Mendlovic and~Raja~Giryes}
\address{School of Electrical Engineering, Faculty of Engineering, Tel-Aviv University}

\begin{document}

\maketitle

\begin{abstract}
Face verification is a fast-growing authentication tool for everyday systems, such as smartphones. While current 2D face recognition methods are very accurate, it has been suggested recently that one may wish to add a 3D sensor to such solutions to make them more reliable and robust to spoofing, e.g., using a 2D print of a person's face.
Yet, this requires an additional relatively expensive depth sensor. To mitigate this, we propose a novel authentication system, based on slim grayscale coded light field imaging. We provide a reconstruction free fast anti-spoofing mechanism, working directly on the coded image. It is followed by a multi-view, multi-modal face verification network that given grayscale data together with a low-res depth map achieves competitive results to the RGB case. We demonstrate the effectiveness of our solution on a simulated 3D (RGBD) version of LFW, which will be made public, and a set of real faces acquired by a light field computational camera. 
\end{abstract}


\section{Introduction}
Automated biometric authentication systems have become increasingly popular in recent years as a mean of identity verification. Among other options, facial recognition is widely used thanks to the abundant labeled facial images available online and to advances in deep learning techniques. In this work, we focus on the task of face-based user authentication. 

Face recognition methods based on RGB data achieve very high accuracy~\cite{DBLP:journals/corr/SchroffKP15,DBLP:journals/corr/abs-1801-09414, Wen2016ADF}. However, while 2D color images are sufficient for face verification, they are insufficient for a complete authentication system, which must be robust to spoofing as one may present a 2D print of a face to such a system. To ensure the authenticity of the user, some existing solutions add a depth sensor based on Time of Flight~\cite{Time-of-Flight} or Structured Light~\cite{iphone_spec} technologies. Compared to the standard 2D setup, the addition of these technologies increases the cost of the authentication system. Therefore, it is of great interest, especially for low-cost devices, to have a system that does not increase the price and  is resilient to 2D spoofing.  

Light field (LF) sensors capture both color and depth information and therefore may be used in face authentication systems. LF images can be captured by various methods such as coded masks, coded apertures, microlenses, and pinhole arrays. Most of these solutions are either impractical or expensive and bulky, as they require a large amount of storage and post-processing time, which make them irrelevant to embedded systems such as smartphones. 

Recently, Marwah\etal have introduced the concept of compressed LF photography~\cite{marwah2013compressive}. They reconstruct a high-resolution LF from its 2D coded projection measured by a single sensor. This allows having a compact system that both provide depth and color information. 

\begin{figure}
    \centering
    \includegraphics[width=0.7\linewidth]{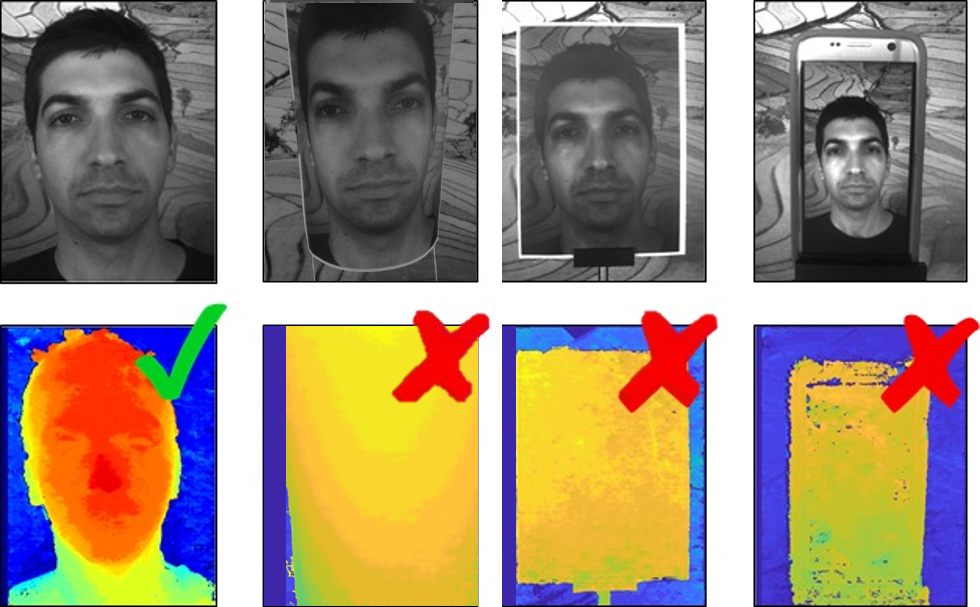}
    \caption{Anti-spoofing demonstrated on images obtained by our light field prototype, rejecting a curved 2D printed image (center left), a flat 2D printed image (center right) and a flat image presented on a smartphone (right). Our anti-spoofing algorithm operates on a coded light field image and does not require the depth images, which are shown here only for demonstration purpose. For details, see Section~\ref{sec:anti_spoof}.}     \label{fig:teaser}
\end{figure}

{\bf Contribution.} In this work, we rely on the concept of compressed LF to introduce a novel low cost face authentication system. It has two major advantages: (i) it allows performing the anti-spoofing operation very efficiently; and (ii) it uses only a grayscale camera with a binary coding mask, which makes the system inexpensive, without reducing the achievable accuracy significantly. 



\section{Coded Light Field Face Authentication}
We now turn to describe our novel grayscale coded LF based authentication system. First, we present the LF coding scheme, which enables fast reconstruction. Then, we show our anti-spoofing mechanism and how we handle face verification on grayscale LF images. 

\subsection{Coded LF Capturing}
We suggest the following simple model of grayscale LF acquisition. For simplicity, we assume the coding of only two views, $view_{0}$ and $view_{1}$, with two effective coding matrices, $\phi_{0}$ and $\phi_{1}$, drawn as different spatial projections of the coding mask $\Phi$. 
Each pixel (at the spatial location $u,v$) in the coded image, $CI$, can, thus, be modeled as
\begin{equation}
\begin{aligned} 
CI(u, v) = \sum_i view_i(u, v)\phi_i(u, v) && i=0, 1.
\end{aligned} 
\end{equation}
Full reconstruction of the views can be obtained using sparse coding or deep learning methods. However, for the task of face authentication, we show that no such reconstruction is needed. The coding mask $\Phi$ may have a random distribution, but it can be measured at system calibration and is therefore known. Using this knowledge, we can obtain a fast, sparse reconstruction of the views by exploiting the mask distribution. For a random binary mask, some sensor pixels code only one view. These pixels' locations can be easily computed by
\begin{equation}
SM_i =  \mathbb{I}[\phi_i > 0] \odot \mathbb{I}[\phi_{1-i}=0],
\end{equation}
where $SM_i$ denotes a "sparse mask" indicating the pixels in which only $view_i$ is captured on the sensor, and $\mathbb{I}$ is the indicator function. 
Therefore, we may obtain a "free" reconstructed sparse view by
\begin{equation}
view_{i, s} = CI \odot SM_i.
\end{equation}
This setup trades resolution for either light efficiency or the number of views being coded. These factors can be optimized to fit a specific application (i.e., more angles can be coded to create a more accurate depth map). Moreover, the coding mask pattern can be optimized given a specific system geometry, which determines the shift between coding matrices.
We focus on the case of only two views. It is possible to extend to more using the same technique. The only constraint is the resolution of the pixels that correspond just to one angle. These can be controlled by changing the the resolution of the captured image and the light efficiency. In our case of two angles, we assume light efficiency of $50\%$, which implies as we describe later that in the coded image quarter of the pixels will belong solely to just one of the angles.
We exploit the fact that face recognition can be done on low-resolution images, so the loss of information in the encoding process might not fail the verification task. 

Our technique is not limited to grayscale images and can be applied also to color images with the appropriate mask or Bayer pattern. Yet, these adjustments complicate the implementation and increase manufacturing cost. Moreover, capturing color information sacrifices resolution and light efficiency, and thus, we focus on the grayscale case showing that depth information can compensate on the absence of color in the face verification task. 

Note that $50\%$ light efficiency means that about a quarter of the pixels in each view can be trivially reconstructed. Assuming a 1.3 MP sensor, of $1080 \times 1400$ resolution, our reconstructed sparse views yield $540 \times 700$ pixels (randomly spaced in the original resolution). Notice that the state-of-the-art RGB face recognition networks receive faces of resolution $250 \times 250$, which may indicate that our ``free'' reconstructions are sufficient for the task of authentication. Our experiments hereafter show that this is indeed the case.

\subsection{Anti Spoofing}\label{sec:anti_spoof}
A reliable face authentication system must ensure that an actual face is being captured, rather than a 2D printed image of a face. In order to do so, other verification systems use expensive 3D scanners such as Time of Flight~\cite{doi:10.1021/jp908381b} or Structured Light~\cite{doi:10.1002/047134608X.W8298} to detect the depth. Besides their cost, notice that the 3D reconstruction process is time consuming. Moreover, some of these systems are power inefficient, sensitive to ambient daylight and only produce a depth signal.
LF images, therefore, hold more promise for the task of face authentication, since they capture both depth and intensity information. 

We suggest the following simple idea, for detecting whether a coded LF image is of a 3D or a 2D object, without the need to fully reconstruct the object. Our method relies on the fact, that the disparity map of a plane being captured in a stereo setting, is also a plane. Let a plane in 3D space be defined by the equation
\begin{equation}\label{eq:1}
c = ax + by + z.
\end{equation}
In a standard stereo setting, the transformation between Euclidean and image spaces is given by
\begin{equation}\label{eq:2}
x = \frac{B}{d}(u-u_0), y = \frac{B}{d}(v-v_0), z = \frac{B}{d}f_u,
\end{equation}
where $B$ is the baseline, $d$ is the disparity measured at the pixel $(u, v)$, $(u_0, v_0)$ is the principle point of the image, and $f_u$ is the pixel focal length. Combining equations~\eqref{eq:1} and~\eqref{eq:2},
\begin{equation}
d = a\frac{B}{c}u + b\frac{B}{c}v + \frac{B}{c}(f_u - au_0 - bv_0).
\end{equation}
Notice that this implies that in the 2D case, the disparity is affine with respect to the image coordinates. Thus, in the case of a flat, 2D object, one can estimate the entire disparity map from a few sampled points. 

This leads us to the following simple anti-spoofing technique. Assume that the captured face is flat. After applying the "free" reconstruction described in the previous section, a basic interpolation can be conducted, followed by a local disparity calculation at three specified image points. The left sparse view can be projected to a new estimation of the right sparse view, by applying the affine disparity map corresponding to the three calculated disparity values, $D_{plane}$:
\begin{equation}
view'_{r, s}(u, v) = view_{l, s}(u + D_{plane}(u, v), v).
\end{equation}
A similarity measure can then be applied on the projected right view $view'_{r, s}$, and the captured sparse right view, $view_{r, s}$. Naturally, we expect this similarity to be lower for captured 3D objects, which have non-flat disparity maps. We found that comparing average $\ell_1$ distance between cubic interpolated sparse images provides good results. 

\subsection{Face Verification}\label{Face Authentication}
As in many other works, we approach the verification task using metric learning. This way, given a fine, accurate metric, we can determine whether two faces belong to the same person or not. We show that combining a multi-view and a multi-modal (having both depth and intensity information) approach, with the power of a large scale pre-trained neural network, enables competitive results in the face verification task on grayscale coded LF data.

Inspired by MVCNN~\cite{Su_multi-viewconvolutional}, which deals with 3D objects shape retrieval rather than faces, we applied a multi-view approach, having different CNN branches learning from 2D projections of a 3D face. Other than the original work, we handle multi-modal data so the weights are not shared and the different input features are concatenated instead of pooled.

The network structure may be summarized as follows. Each input, i.e., left view, right view, and depth map, is fed into an independent Inception Resnet V1~\cite{DBLP:journals/corr/SzegedyIV16} backbone. We choose this network since it has a public pre-trained model available~\cite{davidsandberg_facenet_code}, trained on the 3.3 million faces of VGGFace2~\cite{Cao18}. We then concatenate all the output features and feed them to two fully connected layers. We used triplet loss~\cite{DBLP:journals/corr/SchroffKP15} on the final features to learn the embedding.

\section{Data Generation}\label{Implementaiton}
A major challenge with face verification from coded LF is the amount of needed data: There is no dataset that contains intensity and depth information of the type we acquire with our camera, which is suitable for face recognition, as it requires many different identities. Note that the leading networks of face recognition have been trained on millions of identities and up to two hundred million images~\cite{DBLP:journals/corr/SchroffKP15}.

To evaluate our framework, we generate two types of data: simulated coded 3D data from a 2D dataset and coded data from a prototype we have created. Calibration data from the prototype was utilized in the synthetic dataset creation, in the following manner.

{\bf Grayscale\&D (GD) LFW.} To generate enough data, we use a strategy that creates a 3D face model from its RGB image \cite{Dou2017EndtoEnd3F,Hassner_2013_ICCV,sela2017unrestricted,Sela2017UnrestrictedFG}. We use the published code of~\cite{sela2017unrestricted} to create 3D face reconstructions from the Labeled Faces in the Wild dataset (LFW)~\cite{Huang_labeledfaces}. The 3D model includes a point cloud, a triangulated mesh and a detailed texture. 
Using the relationship between depth and disparity, as measured using our prototype, we convert the point cloud to a disparity map and use it to warp the model into views. This allows transferring networks trained on LFW to the data generated with our system. Also, this gives better disparity maps compared to those calculated from direct point cloud projections. 

{\bf Grayscale LF (GLF) Faces.} In addition, we captured the faces of 88 people using our system prototype, to produce the "GLF Faces" dataset, used for testing our anti-spoofing mechanism and to asses the generalization of the verification network to real data.
Our prototype captured sequential LF, i.e. four separate views (without a coding mask) on a grayscale sensor. We use just two of the captured views, and simulate the coding. 
Other works using such a prototype~\cite{DBLP:journals/corr/abs-1801-10351} show that the simulated coding resembles the physical coding very well. Thus, we believe that the experiment with the real data and simulated coding is a valid proof of concept.

\section{Experiments}\label{Experiments}

\subsection{Anti Spoofing}\label{exp:spoof}
\begin{figure*}[!t]
\centering
\subfloat[Synthetic Data]{\includegraphics[width=0.25\linewidth]{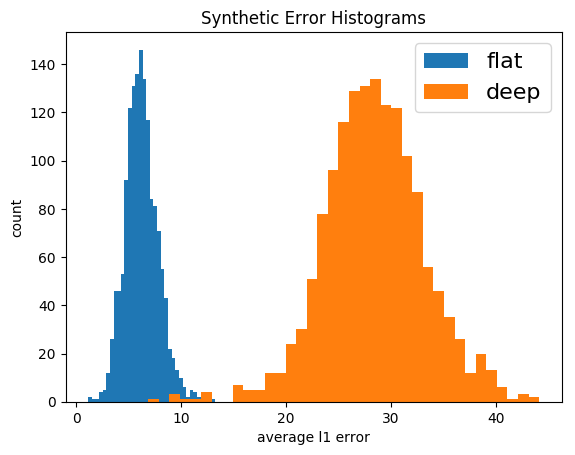}
\label{fig:synth_errors_1494}}
\hfil
\subfloat[Real Data]{\includegraphics[width=0.25\linewidth]{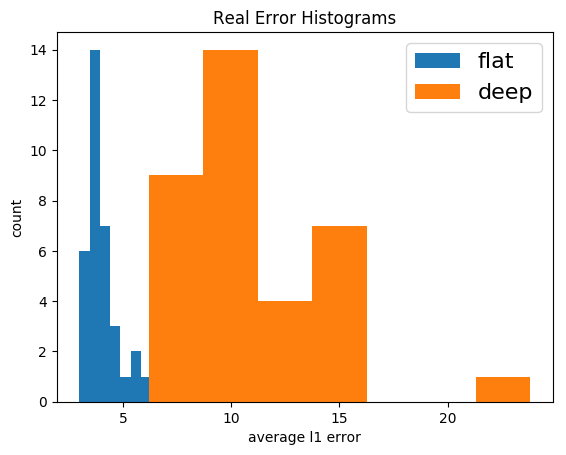}
\label{fig:real_errors}}
\hfil
\subfloat[ROC Curve]{\includegraphics[width=0.25\linewidth]{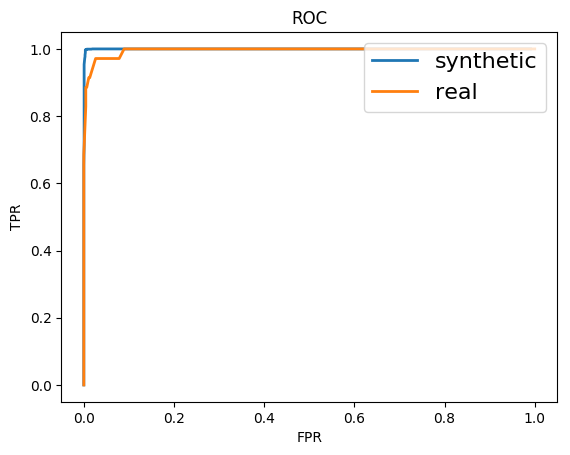}
\label{fig:ROC}}
\caption{{\bf Anti-spoofing results}. \ref{fig:synth_errors_1494}: The average $\ell_1$ error histograms of synthetic projections.
The error in the flat case is generally smaller, indicating that the views are of a printed image of a face. \ref{fig:real_errors}: The average error on our GLF Faces database, including both flat and \textbf{curved} 2D images. \ref{fig:ROC}: The ROC curve in the synthetic (blue) and real (orange) cases.}
\label{fig_sim}
\end{figure*}
To test our anti-spoofing concept, we conduct the following experiment on synthetic data. Using the GD LFW database, which we have created and resembles the depth resolution achieved by our system, we project each grayscale left view to a "flat" right view, by randomized disparity plane parameters. Then, we simulate the acquisition process, resulting in sparse views of both the real and the "flat" projections. Given a sparse left view and a sampled disparity, we create the projected sparse right view, as described in Section~\ref{sec:anti_spoof}. We measure the similarity between the captured (simulated) and projected sparse right views, by the $\ell_1$ loss on their bicubic interpolations. The similarity distributions are plotted in Fig.~\ref{fig:synth_errors_1494}, showing a clear separation of the flat and deep faces.


We repeated the process on the GLF Faces dataset, captured by our system partial prototype. It contains 35 2D views of flat and \textbf{curved surfaces}, such as partial cylinders, and 459 pairs of views of real faces.  We randomly sampled 35 3D views to preserve the scale of the histograms, shown in Fig.~\ref{fig:real_errors}. Though it does not exhibit a clear cut in the separation as in the synthetic case, also in this case, our method can still save a lot of heavy computation time on obvious spoofing cases, which may be the most common type. Notably, the fit error of curved surfaces is still distinctively smaller than that of actual faces, even though they are non-planar. The ROC curves of our anti-spoofing  $\ell_1$-error based classifier are shown in Fig.~\ref{fig:ROC}. Having the verification done also on depth images afterwards prevents the more complicated spoofing scenarios. Yet, it will be applied only on a small fraction of the 2D scans, which were not detected in the first ``cheap'' anti-spoofing test. 

As we are the first to work with coded LF data, we compare our method to the landmarks depth (LDF) method in \cite{LF_PAD_Advanced} which is the-state-of-the-art on the full LF database presented in \cite{LF_PAD_DB}. 
This is the only relevant method for comparison in terms of the data being processed and the computation resources. We report the results using the same terms and measures as in \cite{LF_PAD_DB} on our coded LF dataset. 

We repeated Experiment 2 in \cite{LF_PAD_Advanced}, training for binary classification (genuine or spoofing), mixing all attack types (warped and planar paper, in our case). While keeping the amount of examples of each class equal, we split the data with a $25\%$ test set. We report the average classification error rate (ACER) at the equal error rate (EER), for 50 experiments with random data split. 
Evaluating both strategies on GLF Faces, we outperform their result with ACER of 0.039, compared to 0.08 achieved by LDF. On the synthetic GD LFW, we are also better with $0.004$ ACER compared to $0.05$ achieved by LDF. 
The results are summarized in Table \ref{tab:spoof_comp_results}.
\begin{table}
\centering
\scalebox{0.9}{
 \begin{tabular}{|c c c|} 
     \hline 
      Data & \bf Synthetic & \bf Real\\ [0.5ex] 
     \hline
     LDF & 0.05 & 0.08 \\
     \hline
     Ours & \bf 0.004 & \bf 0.039 \\
     \hline
 \end{tabular}
}
\caption{Average ACER at EER.}
\label{tab:spoof_comp_results}
\end{table}

\subsection{Face Verification}\label{exp:verif}
We report results following the \textit{unrestricted, labeled outside data} protocol of LFW~\cite{Huang2014LabeledFI}, with 10-fold cross-validation. We report a competitive 99.5\% average accuracy on the GD LFW data. For comparison, the original pre-trained network achieves 99.6\% accuracy on LFW, and 81.1\% on grayscale LFW. Fine tuning the pre-trained model (i.e., on a single "branch" of Inception ResNet V1) on a three channel image containing two grayscale views and a depth channel, we got 90\% accuracy, which shows the benefits of our architecture.

On Grayscale LF Faces, the network trained on the synthetic LFW data achieves 91.2\% accuracy on randomly sampled pairs of matching and mismatching identities.
End-to-end fine-tuning on our system data, increases the accuracy to 98.75\%. The test is done in an open set manner, having 78 people in the training set (for the fine tuning) and 10 (different) people in the test set. The same test set is used for both cases (with and without fine-tuning).
Tables~\ref{tab:face_verification_results}, \ref{tab:face_verification_results_real_data} summarize the verification experiments results.
\begin{table}
\centering
\scalebox{0.9}{
 \begin{tabular}{|c c c|} 
 \hline 
  & \bf Accuracy \% & \bf Comments\\ [0.5ex] 
 \hline
 FaceNet & 99.6 & RGB images; 10 folds \\
 \hline
 FaceNet & 90 & finetuned on GD images; 1 fold \\
 \hline
 Ours & 99.5 & GD images; 10 folds\\
 \hline
\end{tabular}
}
\caption{Average verification accuracy on GD LFW.}
\label{tab:face_verification_results} 
\end{table}

\begin{table}
\centering
\scalebox{0.9}{
 \begin{tabular}{|c c|} 
 \hline 
  Network & \bf Accuracy\\ [0.5ex] 
 \hline
 Transferred & 91.2\\
 \hline
 Fine-tuned & 98.75\\
 \hline
\end{tabular}
}
\caption{Average verification accuracy on our GLF dataset.}
\label{tab:face_verification_results_real_data} 
\end{table}

\section{Conclusion}\label{Conclusion}
To conclude, this work presents a novel face authentication system based on grayscale coded LF, eliminating the need for a complicated and possibly expensive reconstruction flow. Our proof of concept includes a fast and simple anti-spoofing mechanism that works directly on the coded image. We achieved competitive results in the face verification task based on the coded data, showing a successful manifestation of the concept of application optimized sensing systems. Printing a 2D image is quite an easy fraud, so being able to detect it in an effective way is useful for real-life products. 

Note that the focus of this work is to provide a reliable low-cost system for face authentication and not to overcome all possible adversaries. 
All anti-spoofing methods can be tricked, given enough effort. Clearly, if one would create a 3D model from a person's 3D scan, it will fool our system, and most of the other 3D based technologies as well. Yet, a video can be combined with our solution, to overcome such challenging spoofing attempts. We leave this to future work.
%


\bibliographystyle{IEEEbib}
\bibliography{egbib}

\end{document}